\documentclass[sigconf]{acmart}

\AtBeginDocument{%
  \providecommand\BibTeX{{%
    \normalfont B\kern-0.5em{\scshape i\kern-0.25em b}\kern-0.8em\TeX}}}

 \setcopyright{none}

\usepackage[inline]{enumitem}
\usepackage{makecell}
\usepackage{fancyhdr}

\acmConference[Proceedings of the Data Collection, Curation, and Labeling for Mining and Learning]{Proceedings of the Data Collection, Curation, and Labeling for Mining and Learning}{August 05, 2019}{Anchorage, Alaska, USA}

\begin{document}
\fancyhead{}
\fancyfoot{}
\title{Automated Discovery and Classification of Training Videos for Career Progression}


\author{Alan Chern, Phuong Hoang, Madhav Sigdel, Janani Balaji, and Mohammed Korayem}

\affiliation{%
  \institution{CareerBuilder LLC}
  \streetaddress{5550 Peachtree Parkway}
  \city{Greater Atlanta Area}
  \state{Georgia, US}
  \postcode{43017-6221}
}
\email{{alan.chern, phuong.hoang, madhav.sigdel, janani.balaji, mohammed.korayem}@careerbuilder.com}

\renewcommand{\shortauthors}{Chern and Hoang, et al.}

\begin{abstract}
Job transitions and upskilling are common actions taken by many industry working professionals throughout their career. With the current rapidly changing job landscape where requirements are constantly changing and industry sectors are emerging, it is especially difficult to plan and navigate a predetermined career path. In this work, we implemented a system to automate the collection and classification of training videos to help job seekers identify and acquire the skills necessary to transition to the next step in their career. We extracted educational videos and built a machine learning classifier to predict video relevancy. This system allows us to discover relevant videos at a large scale for job title-skill pairs. Our experiments show significant improvements in the model performance by incorporating embedding vectors associated with the video attributes. Additionally, we evaluated the optimal probability threshold to extract as many videos as possible with minimal false positive rate.
\end{abstract}




\maketitle

\section{Introduction}
Career progression is a very important part of the career plans and goals of many working professionals. However, there are generally some hardships when attempting to transition to a new position in the dynamic job market today where new job sectors are constantly emerging with changing technologies and requirements. 
Job seekers are often required to acquire the skills necessary for the new position (i.e., upskilling). It is difficult to identify the most important set of skills among various job postings between employers, and further searching for suitable resources to learn these skills poses an additional challenge.

Many researchers and organizations have proposed and implemented various ways to model career paths and analyze the associated skill gaps. Some works mined professional profiles to model the similarity between career paths \cite{xu2014modeling}, analyze job-hopping behavior \cite{oentaryo2017analyzing}, and predict career transitions via career network model \cite{safavi2018career} and contextual embedding \cite{li2017nemo}. Xu et al. built job transition networks to detect talent circles for talent recommendation \cite{xu2016talent}. Wowczko analyzed skills demand based on job vacancy data \cite{wowczko2015skills}. However, it is still ultimately up to the job seeker to look for the appropriate resources to attain the skills they require.

\begin{figure}[t]
    \centering
    \includegraphics[width=\linewidth]{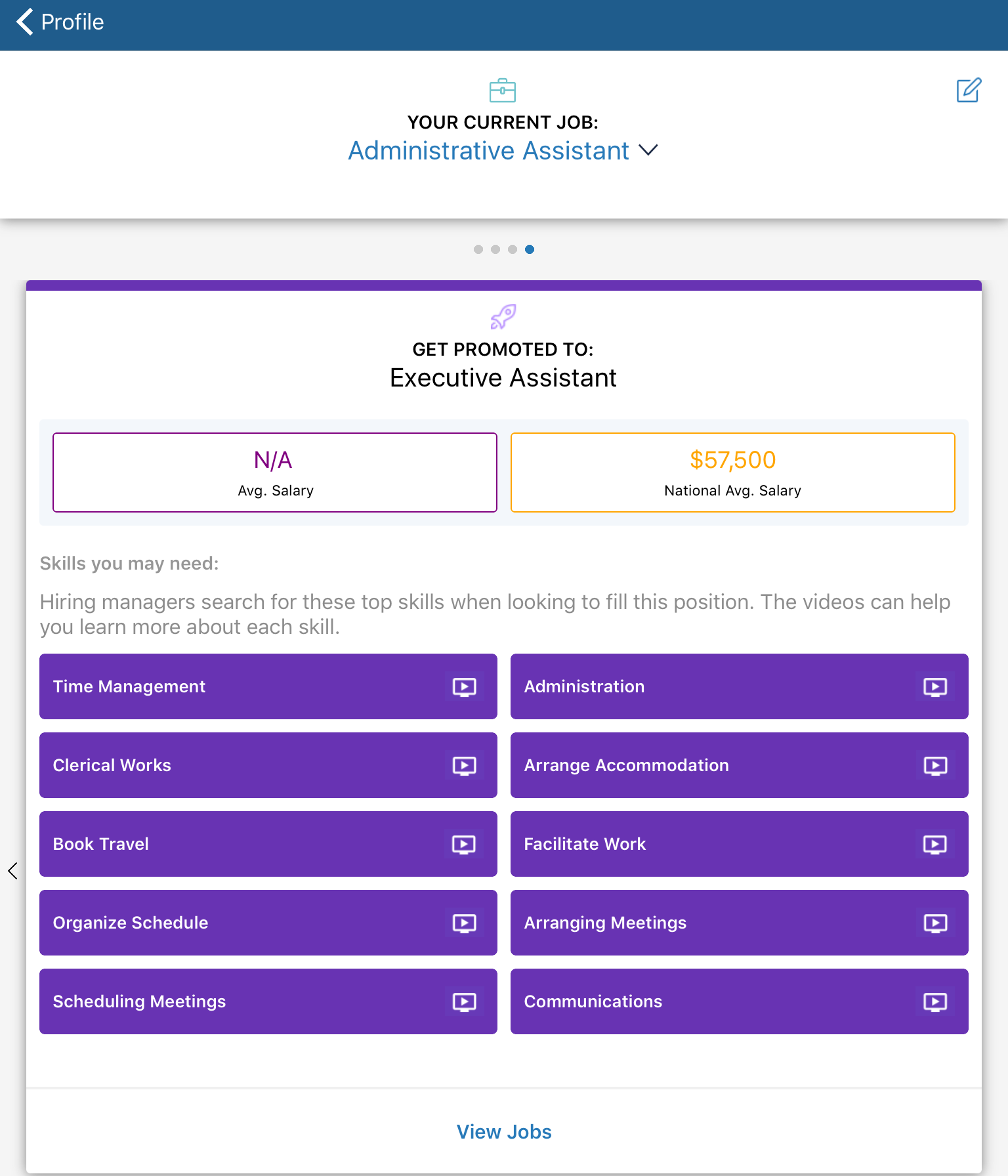}
    \caption{{\itshape Next Job} mobile user interface. One of the recommended job transition from {\itshape Administrative Assistant} is {\itshape Executive Assistant}.}
    \label{fig:nextjob_ui}
    \Description{{\itshape Next Job} mobile user interface.}
\end{figure}

\begin{figure}[t]
    \centering
    \includegraphics[width=\linewidth]{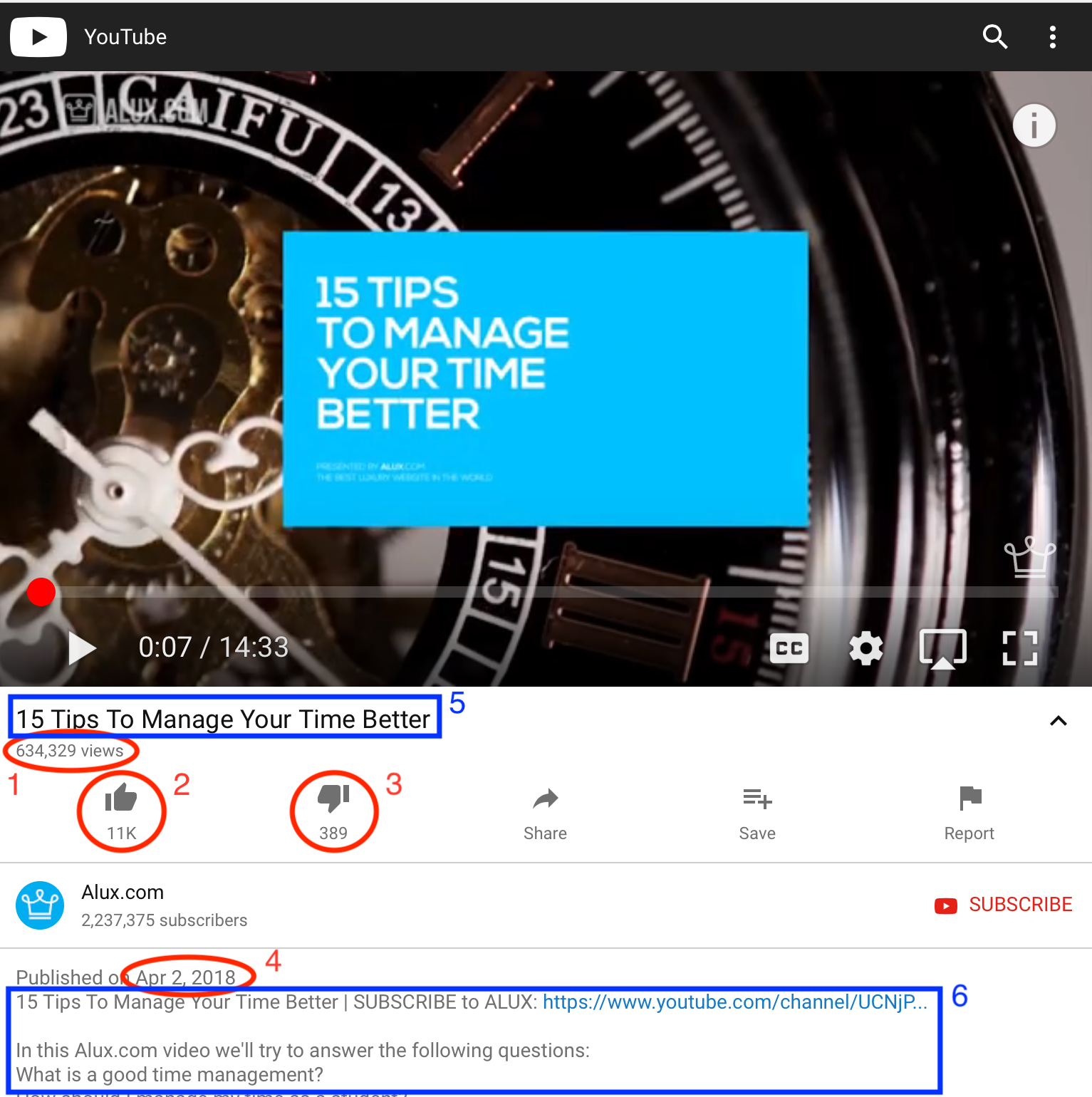}
    \caption{Example YouTube educational video snapshot for the ``Time Management'' skill. The video provides tips for viewers to better manage time. The numerical labels denote some of the signals used as features to train the machine learning classifier.}
    \label{fig:nextjob_youtube}
    \Description{Example YouTube educational video snapshot for the ``Time Management'' skill.}
\end{figure}

CareerBuilder is a human capital management company with over 90 million users globally. It provides various products and services to assist users to find jobs. One of the products is {\itshape Next Job}, which is designed to guide users through potential career paths. This is achieved by our own job transition and job-skill networks modeled using a combined representation learning approach \cite{dave2018combined}. However, most importantly, the main focus of this work is that in addition to suggesting job titles and the related skills, we are incorporating the means to acquire these skills. Specifically, given a job title, {\itshape Next Job} provides the recommended job transition(s), the most critical skills needed for each new job title, and the resources to obtain these skills, if available. For instance, the {\itshape Administrative Assistant} can advance its career to an {\itshape Administrative Coordinator} or an {\itshape Executive Assistant} (shown in Figure~\ref{fig:nextjob_ui}). The additional skills that may be needed for an {\itshape Executive Assistant} include ``Time Management,'' ``Administration,'' ``Clerical Works,'' etc. By clicking on these skills, the application is directed to a public educational video (hosted on YouTube) that aims to either teach or inform the user about this particular skill. An example snapshot of the video corresponding to the ``Time Management'' skill can be seen in Figure~\ref{fig:nextjob_youtube}, which provides tips for viewers to better manage time.

While there are an abundance and a wide variety of learning resources and materials publicly available, we believe that videos tend to be the most engaging option. Furthermore, even though many online platforms offer educational videos, they are often not free of cost and/or are limited to specialized topics. As a result, we selected YouTube because of the breadth as well as the depth of educational video contents it offers at no cost to the users.

The main challenges we faced when building this product are:
\begin{enumerate}
    \item Finding the appropriate video for a title-skill pair with acceptable video and audio quality. It is difficult to determine the subset out of all available videos that are suitable for the end-user.
    \item Scaling this search for combinations of thousands of unique job titles coupled with tens of distinct skills per title. It is important to note that different job titles may share the same skills, but a video considered suitable for a certain job title-skill pair does not guarantee its usability for the same skill paired with another job title.
\end{enumerate}

These tasks require careful data collection, curation, and labeling, all of which can be both expensive and time-consuming. Therefore, we designed and implemented a machine learning based system to automate our video collection and classification.

\section{End-to-End System Flow}
\begin{figure}[t]
    \centering
    \includegraphics[width=\linewidth]{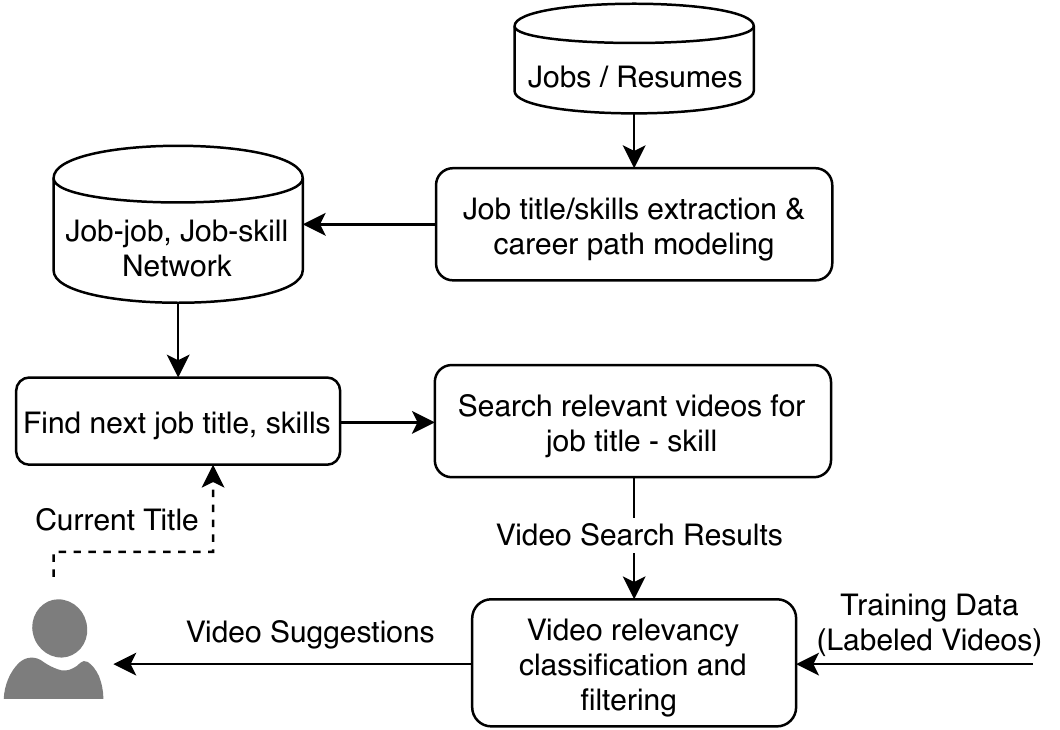}
    \caption{Automated System Flow}
    \label{fig:nextjob_system}
    \Description{Automated System Flow}
\end{figure}

Our system is outlined in Figure~\ref{fig:nextjob_system}, and begins with the extraction of the raw jobs and resumes data that we have collected from employers and job seekers. These data are parsed and normalized for training our career path model \cite{dave2018combined}. This model consists of a job-job and a job-skill network that provide the next job along with associated skills given a current job title. In total, we have 5,425 unique normalized job titles \cite{Javed2015} and an average of 15 normalized skill terms \cite{Hoang2018} per title.

We use the YouTube API \footnote{https://developers.google.com/youtube/v3/} to retrieve the videos by feeding in our customized search queries. For each job title-skill pair, the query gets formed in three different ways to try to capture all related videos. Specifically, the queries are generated by
\begin{enumerate*}
    \item skill term
    \item skill term + job title
    \item ``skill term'' + job title
\end{enumerate*} \footnote{Quotation indicates the exact skill term must appear in the search results.}.
The API calls are configured to fetch only video contents in English. Additionally, based on manual assessment, we find that limiting the video category to ``Education'' tends to yield results more relevant for our purpose. Next, we de-duplicate the extracted videos and repeat the retrieval step until we get up to nine unique videos per title-skill pair. If YouTube does not return enough results, we retain what we have and continue on.

For training our classifier, we first select around 350 titles and their associated skills. Then, our internal data quality team manually curated and assigned a binary label to each video (i.e., either a positive label for relevant or negative for irrelevant). The relevancy is decided solely based on the video content and irrespective of any video viewer statistics. This provided us with around 48,000 labeled videos to use for our model training. The distribution of positive and negative labels is roughly even with approximately 43\% being positive. We select the Random Forest algorithm to perform our video relevance classification. Details of the model training and evaluation processes will be discussed in Section~\ref{sec:experiments}.

Finally, the classifier is used to automatically categorize the retrieved videos for the rest of our title and skill pool. As we maximize the precision of our model, we are able to confidently rely on it to determine whether a video is relevant or not. The videos marked as irrelevant get discarded and the remaining relevant videos can be displayed to the user.



\section{Experiments and Results}
\label{sec:experiments}
After the manual curation and labeling, the labeled video data set is split into training and testing sets with a ratio of 80:20 to respectively train and evaluate our classifier. Note that the distribution of positive and negative labels are roughly equal, with the positive (i.e., ``relevant'') being slightly less. We select the Random Forest algorithm to train our training set and tune the hyperparameters via cross validation. Then, we compare the performance on the testing set with two different sets of features used to train our model. Finally, the model utility precision and false positive rate are evaluated based on the number of title-skill pairs it successfully retrieves at least one relevant video for.

\subsection{Training Features}
\label{sec:training_features}
Table~\ref{tab:features} summarizes the features and their membership in the first feature set (namely, ``feature set 1'') and the second feature set (namely, ``feature set 2''). The reference numbers correspond to the numerical labels shown in Figure~\ref{fig:nextjob_youtube}, illustrating some of the signals that are part of YouTube videos which we are able to retrieve and use for training our model. Note that a few of the features we used are not displayed on the graphical user interface but are nevertheless available via the API.

Feature set 1 consists of only the statistics related to the video that can be extracted using the YouTube API, including the view, like, dislike, comment counts and the ratio between each count pair, video duration, and the number of days the video has been published for. Feature set 2 includes all the features from the first with the addition of embedding vectors \cite{le2014distributed} associated with the video title and description. We also incorporate three embedding vectors for the search query, job title, and skill (i.e., ``Other vectors'') as features. The embedding vectors used here are each 512-dimensional and generated using a pre-trained deep learning model that was built with our internal resume and job data and converts input text into the embedding vector space. Lastly, the cosine similarities between each of these vector pairs are added to the feature set.

Because precision is the most important metric to our system, we tune the model to maximize this metric. Table~\ref{tab:metrics} shows the results for the precision, recall and f-score of each set of features used.

\begin{table}[t]
  \caption{Training Feature Sets}
  \label{tab:features}
  \begin{tabular}{lccc}
    \toprule
    Feature & Ref. no. & Feature set 1 & Feature set 2 \\
    \midrule
    View count & 1 & \checkmark & \checkmark \\
    Like count & 2 & \checkmark & \checkmark \\
    Dislike count & 3 & \checkmark & \checkmark \\
    Comment count & n/a & \checkmark & \checkmark \\
    Count ratios & n/a & \checkmark & \checkmark \\
    Days elapsed & 4 & \checkmark & \checkmark \\
    Title (vector) & 5 & & \checkmark \\
    Description (vector) & 6 & & \checkmark \\
    Other vectors & n/a & & \checkmark \\
    Vector similarities & n/a & & \checkmark \\
    \bottomrule
  \end{tabular}
\end{table}

\begin{table}[h]
  \caption{Result Metrics of Feature Sets}
  \label{tab:metrics}
  \begin{tabular}{lccc}
    \toprule
    & Precision & Recall & F-Score \\
    \midrule
    Feature set 1 & 0.688 & 0.621 & 0.653\\
    Feature set 2 & \textbf{0.767} & \textbf{0.645} & \textbf{0.701}\\
    \bottomrule
  \end{tabular}
\end{table}

\subsection{Optimal Probability Threshold}
\label{sec:probability_threshold}
We use feature set 2 to further evaluate our model. Specifically, instead of using the model to predict a label directly (i.e., a probability threshold of 0.5), we let it output the probability of the label being positive. Then, we vary the positive probability thresholds over a range to compare the utility precision ($P_{u}$) and false positive rate ($FPR$). The utility precision is defined as
$$P_{u} = \frac{N_{\mbox{predicted}}}{N_{\mbox{labeled}}}$$
where the $N_{\mbox{predicted}}$ is the count of unique title-skill pairs with at least one video correctly predicted as relevant, and $N_{\mbox{labeled}}$ is the total number of distinct title-skill pairs with at least one video manually labeled as relevant. The false positive rate is defined as
$$FPR = \frac{FP}{FP+TN}$$
where $FP$ is the number of false negatives (i.e., incorrectly predicted positive labels), and $TN$ is the number of true negatives. As such, the denominator of $FP+TN$ denotes the total number of actual negative labels.

\begin{figure}[h]
    \centering
    \includegraphics[width=\linewidth]{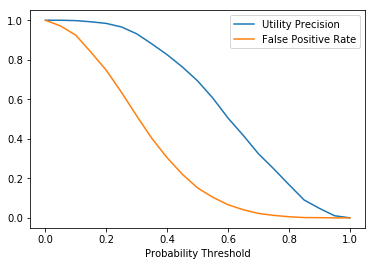}
    \caption{Utility precision and false positive rate vs. the model probability threshold. A low probability threshold yields a high utility precision but also a high false positive rate, while a high probability threshold results in low values of both utility precision and false positive rate.}
    \label{fig:precision}
    \Description{Utility precision and false positive rate vs. the model probability threshold.}
\end{figure}

Figure~\ref{fig:precision} shows the trade-off between the utility precision and the false positive rate as the probability threshold for our model varies. Setting too low of a probability threshold means that more relevant videos would be captured overall, but would also lead to higher chances of incorrectly marking an irrelevant video as relevant. The opposite trend would be exhibited by setting the probability threshold too high. Thus, we need to find a good balance between being able to obtain at least one relevant video per title-skill pair while maintaining a low number of wrongly categorized irrelevant videos.

\begin{table*}[t]
  \caption{Video Feature Subset Samples}
  \label{tab:feature_samples}
  \begin{tabular}{ccccccccc}
    \toprule
    \textbf{Job title} & \textbf{Skill} & \textbf{Video title} & \makecell{\textbf{Video description} \\ \textbf{(partial)}} & \makecell{\textbf{Skill-video} \\ \textbf{title similarity}} & \makecell{\textbf{Job title-video} \\ \textbf{desc. similarity}} & \makecell{\textbf{View} \\ \textbf{count}} & \makecell{\textbf{Like} \\ \textbf{count}} & \makecell{\textbf{True} \\ \textbf{label}} \\
    \toprule
    Recruiter & \makecell{Interview \\ Scheduling} & \makecell{Scheduling \\ an Interview} & \makecell{Step-by-step tutorial \\ on how to schedule \\ an interview...} & 0.858 & 0.551 & 126 & 1 & + \\
    \midrule
    \makecell{Administrative \\ Specialist} & Spreadsheets & \makecell{5 Excel \\ Questions \\ Asked in Job \\ Interviews} & \makecell{Top 5 Excel \\ Interview Questions. \\ These MS Excel \\ interview questions \\ and answers...} & 0.028 & 0.229 & 1.1M+ & 18K+ & - \\
    \bottomrule
  \end{tabular}
\end{table*}

\section{Discussion}
The results shown in Section~\ref{sec:training_features} demonstrate that the model performance is better when including the additional embedding vectors as training features. The most important metric (precision) value of 0.767 significantly outperforms the feature set without the embedding vectors. This can be explained by the fact that relevant videos tend to possess titles and descriptions which are helpful in determining its relevance. Thus, these vectors enable the model to better classify relevant videos that may be newer (i.e., have none or very low statistics counts) and irrelevant videos that may have favorable statistics. An example of each of these scenarios is shown in Table~\ref{tab:feature_samples}. For the job title of ``Recruiter,'' an actually relevant video with low statistics is classified as irrelevant by the model when using feature set 1, but because feature set 2 takes into account the high vector cosine similarity values between the different attributes, it is able to correctly predict its label. In contrast, for the ``Administrative Specialist,'' while the associated video possesses very high statistics and gets incorrectly labeled by the model with feature set 1, using feature set 2 allows the model to again consider the cosine similarities, which are relatively very low, between the video attributes, enabling it to predict the correct positive label.

Therefore, by incorporating the embedding vectors for the various video attributes and the job title and skill, as well as the cosine similarity values between each vector pair, the model is able to better utilize all the available signals, which results in better performance.

Additionally, our evaluation of the utility precision and false positive rate in Section~\ref{sec:probability_threshold} demonstrate that a probability threshold of 0.6 yields the optimal balance between the number of relevant videos retrieved and the number of false positives. As can be seen in Figure~\ref{fig:precision}, as this point, the utility precision is relatively high around 0.5 while the false positive rate starts to flatten out to below 0.07. These results signify that our system is able to automatically select a relevant video for over half of the title-skill pairs with less than 7\% error rate.

\section{Conclusion and Future Work}
The {\itshape Next Job} product at CareerBuilder is designed to recommend suitable career paths for users and assist them in upskilling and preparing for their next job. For a given job title, the user is shown at least one potential job progression along with the skills that may be necessary to achieve such transitions. Each skill is associated with an education video, if available, to help the user learn and attain the skill.

As we have over 5,000 distinct normalized job titles, each with an average of over 15 recommended skills, there are two major challenges that we have to overcome. The first is to be able to find appropriate videos for each of the title-skill pairs, and the second is to scale this video retrieval for all such pairs. Therefore, we designed a data collection and classification system to automatically extract relevant videos for each title-skill pair.

The evaluation results for our system show that the machine learning classifier performance is significantly improved by incorporating the embedding vectors associated with the video title and description, and the search query, job title, and skill as features along with the video statistics. These additional features allow the model to take more useful signals into consideration. Furthermore, we found that by adjusting the probability threshold of the model for positive label, the system can achieve an optimal balance between utility precision and false positive rate. This allows an automated retrieval of relevant videos for over 50\% of the title-skill pairs with a false positive rate of less than 7\%.

A limitation of our proposed system is that it only performs binary classification (i.e., relevant or irrelevant) on training videos for a given job title-skill pair. In practice, the model would fall short in discovering training videos suitable for users having different skill proficiency levels (e.g., beginner, intermediate, advanced). Secondly, it would be interesting to track user interactions toward the suggested videos and integrate these feedback signals to improve our recommendations; for instance, replacing videos that users dislike (skip or stop early on). A reinforcement learning framework would be a suitable technique to overcome this limitation. We leave the development of such a framework as our future work.




\bibliographystyle{ACM-Reference-Format}
\bibliography{references}

\end{document}